\documentclass[runningheads]{llncs}
\usepackage[T1]{fontenc}
\usepackage{multicol,multirow}
\usepackage{graphicx,verbatim}
\usepackage{amssymb}

\begin{document}
\title{Evaluating Foundation Models with Pathological Concept Learning for Kidney Cancer}
\author{Shangqi Gao\inst{1}\thanks{Corresponding author: sg2162@cam.ac.uk} \and
Sihan Wang\inst{2} \and
Yibo Gao\inst{2} \and Boming Wang\inst{2} \and Xiahai Zhuang\inst{2} \and Anne Warren\inst{1,3} \and Grant Stewart\inst{1,3} \and James Jones\inst{1,3} \and Mireia Crispin-Ortuzar\inst{1}}
%
\authorrunning{S. Gao et al.}
%
\institute{University of Cambridge, Cambridge, UK \and Fudan University, Shanghai, China \and 
Cambridge University Hospitals NHS Foundation Trust, Cambridge, UK
}


\maketitle              
\begin{abstract}
To evaluate the translational capabilities of foundation models, we develop a pathological concept learning approach focused on kidney cancer. By leveraging TNM staging guidelines and pathology reports, we build comprehensive pathological concepts for kidney cancer. Then, we extract deep features from whole slide images using foundation models, construct pathological graphs to capture spatial correlations, and trained graph neural networks to identify these concepts. Finally, we demonstrate the effectiveness of this approach in kidney cancer survival analysis, highlighting its explainability and fairness in identifying low- and high-risk patients. The source code has been released by \url{https://github.com/shangqigao/RadioPath}.


\end{abstract}
\section{Introduction}
Kidney cancer is the 14th most common cancer and the 16th leading cause of cancer death globally, with renal cell carcinoma (RCC) being the most common type \cite{intro_kidneycaner}. RCC subtyping is critical for treatment, as different subtypes have distinct behaviors and survival rates. Besides, early detection is vital for prognosis, but 42\% of cases in UK are diagnosed at advanced stages \cite{intro_ERCD,intro_kidneyreview}. Foundation models have significantly enhanced the ability of computational pathology frameworks to make clinically relevant predictions~\cite{intro_UNI,intro_CONCH,intro_CHIEF}. However, the black-box nature of the downstream, task-specific decision-making models limits their ability to provide actionable insights for pathologists. Enhancing AI model transparency is essential for clinical translation~\cite{intro_interpretability_medicalAI}, which is in line with recent U.S. FDA guidelines on machine learning-enabled medical devices \cite{intro_FDA}.


Recent Concept Bottleneck Models (CBMs) \cite{intro_CBM} provide an interpretable approach by incorporating human-interpretable concepts into the decision-making process, enabling AI models to recognize and utilize medically meaningful intermediate features. CBMs break tasks into two stages: predicting concepts from images and using those concepts for task prediction. These models have been successfully applied in medical domains like knee X-ray grading~\cite{intro_CBM} and skin disease diagnosis~\cite{evi_cem}. In pathology, CBMs are promising in identifying pathological concepts, from fine cellular features to coarse histopathological patterns.

Building on the success of Concept Bottleneck Models (CBMs), we develop pathological concept learning for kidney cancer. Our contributions are as follows,
\begin{itemize}
    \item We develop pathological concept learning for kidney cancer by creating a comprehensive set of kidney-related pathological concepts derived from the TNM staging guidelines and pathology reports.
    \item We explore the potential of foundation models in pathology by training graph neural networks (GNNs) on constructed pathological graphs to evaluate their ability to identify pathological concepts.
    \item We demonstrate the explainability of pathological concept learning in uncovering risk factors for kidney cancer survival and validate its fairness in identifying low- and high-risk patients.
\end{itemize}

\section{Methodology}

\begin{figure}
    \centering
    \includegraphics[width=1\linewidth]{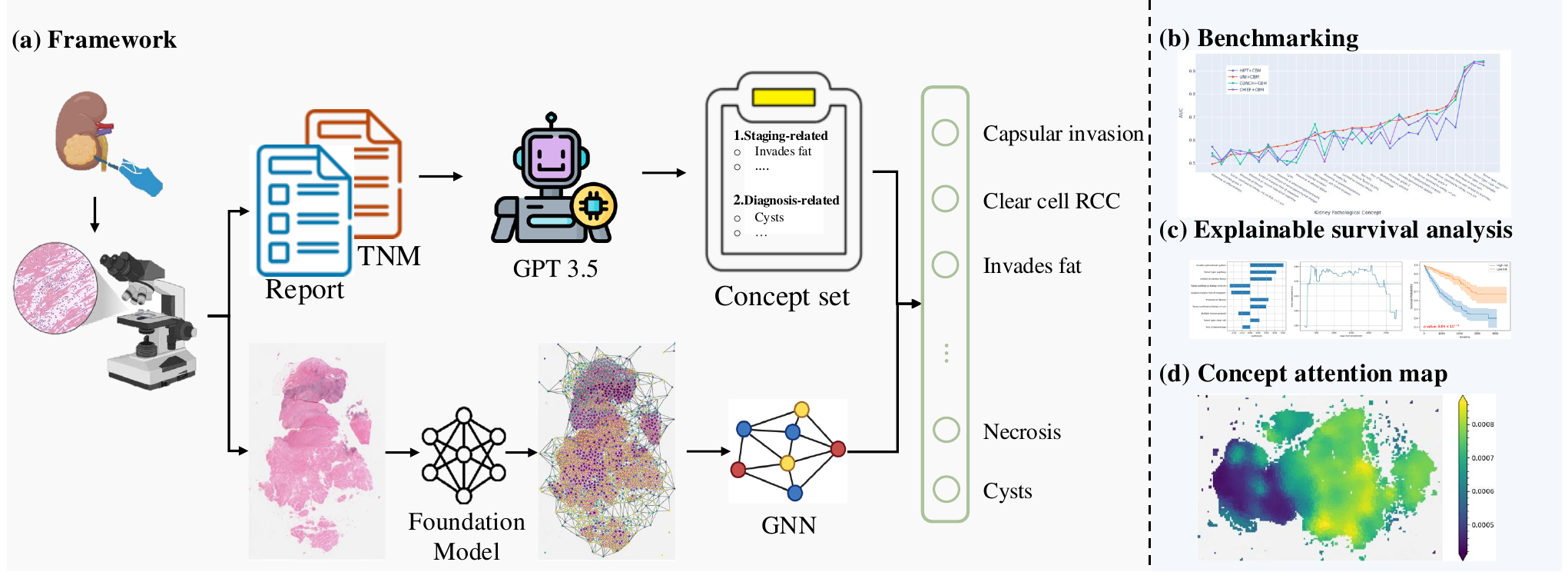}
    \caption{Overview of pathological concept learning. (a) The framework of pathological concept learning; (b) benchmarking foundation models in identifying pathological concepts; (c) Explainable survival analysis based on pathological concepts; and (d) Identification of spatial phenotypes by concept-orientated attention map.}
    \label{fig:pipeline}
\end{figure}

\textbf{Construction of kidney pathology concepts.} For pathological concept learning, we first construct a comprehensive concept set guided by clinical standards, as shown in Fig. \ref{fig:pipeline}. This set is built by extracting concepts from two sources: the TNM staging guidelines for renal cancer \cite{TNM}, which provide tumor-specific attributes like size and invasion, and pathology reports, from which we use GPT-3.5 to identify concepts related to renal lesions. The concept set is designed to offer both local and global perspectives, such as "necrosis" for localized features and "invasion beyond Gerota’s fascia" for broader context. These diverse concepts present challenges for contemporary foundation models, as discussed in Sections \ref{experiment:concept learning} and \ref{experiment:survival analysis}.

\textbf{Identification of pathological concepts.} In the second step of pathological concept learning, we identify diverse concepts by learning spatial relationships between different regions in whole slide images (WSIs). We first extract image features patch-by-patch using pathology foundation models and construct whole slide graphs \cite{bladder_BGNNs}. A graph neural network (GNN) with Linear, ReLU, and GATConv layers is then used to learn spatial tissue structures and reduce feature dimensions to match the number of pathological concepts. Spatial features are aggregated using an attention-based multiple instance learning (ABMIL) \cite{ABMIL} widely adopted in foundation models \cite{intro_UNI,intro_CHIEF}, where each concept has its own attention network to capture specific spatial responses. The network is trained by minimizing binary cross-entropy loss for each concept, as follows,
\begin{equation}
    \min_{\theta} \sum_{i=1}^N \mathrm{BCE}(f_{\theta}(G_i), y_i),
\end{equation}
where, $G_i=(V_i, E_i)$ denotes a WSI graph with vertex set $V_i$ and edge set $E_i$, $f_{\theta}(\cdot)$ represents the graph neural network parameterized by $\theta$ for pathological concept learning, $y_i\in \mathbb{R}^k$ denotes the label of $K$ concepts, and $N$ denotes the number of WSIs for training.

\textbf{Explainable survival analysis.} Conventional computational pathology models predict survival based on aggregated hidden features, which do not provide interpretable risk factors. In contrast, the explainability of pathological concepts allows for a fully transparent decision-making process in survival analysis. Specifically, we use the Cox proportional hazards (CoxPH) model to predict survival based on the concepts identified through our pathological concept learning. This approach enables us to pinpoint high-risk pathological concepts for mortality and assess the importance of each concept in survival analysis.

\section{Experiments}
\subsection{Datasets and implementation details}
\textbf{Datasets.} The TCGA-RCC dataset, comprising TCGA-KIRC, TCGA-KIRP, and TCGA-KICH, includes 947 subjects with WSIs and pathology reports, and 941 subjects with clinical outcomes \footnote{https://www.cancer.gov/ccg/research/genome-sequencing/tcga}. For graph construction, we successfully built 944 graphs from the 947 WSIs. In concept extraction, GPT-3.5 successfully processed 776 out of 947 pathology reports. After matching with the 941 clinical outcomes, we obtained 775 samples with WSI graphs, concept labels, and clinical outcomes for training and testing, as summarized in Table \ref{tab:demographics}. Finally, we applied 5-fold cross-validation, splitting the 775 samples into 620 for training and 155 for testing with an 8:2 ratio.

\textbf{Metrics.} To comprehensively evaluate pathological concept learning and RCC subtyping, we used four widely adopted metrics: balanced accuracy (ACC), F1 score, Area Under the Curve (AUC), and average precision (AP). For survival analysis, we assessed performance using the Concordance Index (C-Index), the Concordance Index with inverse probability of censoring weights (C-IPCW), cumulative/dynamic AUC (C-AUC), and the Integrated Brier Score (IBS).

\textbf{Implementation.} Before feature extraction, we applied the OTSU tissue masking method to select WSI regions containing tissue samples. For WSI patch embedding, we used four foundation models—HIPT \cite{intro_HIPT}, UNI \cite{intro_UNI}, CONCH \cite{intro_CONCH}, and CHIEF \cite{intro_CHIEF}—to extract spatial features patch-by-patch. Training was conducted using the ADAM optimizer with an initial learning rate of $3\times 10^{-4}$ and a weight decay of 
$1\times 10^{-5}$. The total training steps were set to 20, with the learning rate following a cosine annealing schedule. All models were trained and tested on an NVIDIA L40 GPU with a batch size of 32.

\subsection{Concept learning} \label{experiment:concept learning}

\begin{table}[!t]
    \centering
    \begin{tabular}{c|c|c|c|c|c|c|c|c|c|c}
        \hline
        Age & \multicolumn{2}{c|}{Gender} & \multicolumn{5}{c|}{Race} & \multicolumn{3}{c}{Subtype} \\
        \cline{2-11}
         $Mean_{std}$& Female & Male &  White&  Black&  Asian&  Unknown&  N/A&  ccRCC&  pRCC&  chRCC\\
         \hline
         $59.7_{12.7}$&  257&  518&  641&  107&  13&  13&  1&  480&  217&  78\\
         \hline
    \end{tabular}
    \caption{Summary of patient demographics for our study}
    \label{tab:demographics}
\end{table}

\begin{table}[!tb]
    \centering
    \begin{tabular}{c|c|c|c|c|c|c|c|c}
        \hline
        \multirow{2}{*}{Model} & \multicolumn{2}{c|}{ACC} & \multicolumn{2}{c|}{F1 Score}& \multicolumn{2}{c|}{AUC} & \multicolumn{2}{c}{AP} \\
        \cline{2-9}
        & Top10& Mean& Top10& Mean& Top10& Mean& Top10& Mean \\
        \hline
        HIPT+CBM & 0.620& 0.540& 0.675& 0.255& 0.735& 0.626& 0.738& 0.438 \\
        UNI+CBM & \textbf{0.682}& \textbf{0.564}& \textbf{0.758}& \textbf{0.316}& \textbf{0.789}& \textbf{0.661}& \textbf{0.784}& \textbf{0.476} \\
        CONCH+CBM & \textit{0.660}& \textit{0.555}& \textit{0.743}& \textit{0.293}& \textit{0.782}& \textit{0.642}& \textit{0.776}& \textit{0.459} \\
        CHIEF+CBM & 0.648& 0.551& 0.719& 0.285& 0.777& 0.639& 0.760& 0.455 \\
        \hline
    \end{tabular}
    \caption{Pathological concept learning based on different foundation models.}
    \label{tab:concept learning}
\end{table}

In this section, we evaluated the overall performance of concept learning and provided a detailed analysis of its effectiveness in cancer subtyping. For WSI feature extractors, we used four foundation models, i.e., HIPT, UNI, CONCH, and CHIEF. For convenience, we denote each model as "extractor + CBM" (e.g., "HIPT + CBM").

\textbf{Overall
concept learning performance.}
Table \ref{tab:concept learning} summarizes the results of concept learning. Among the evaluated models, UNI + CBM achieved the highest performance across all metrics, with Top-10 ACC, F1 score, AUC, and AP values of 0.682, 0.758, 0.789, and 0.784, respectively. CONCH + CBM ranked second, delivering competitive performance with corresponding scores of 0.660, 0.743, 0.782, and 0.776. These results indicate that UNI outperforms other models in feature extraction.

\begin{figure}[!tb]
    \centering
    \includegraphics[width=1\linewidth]{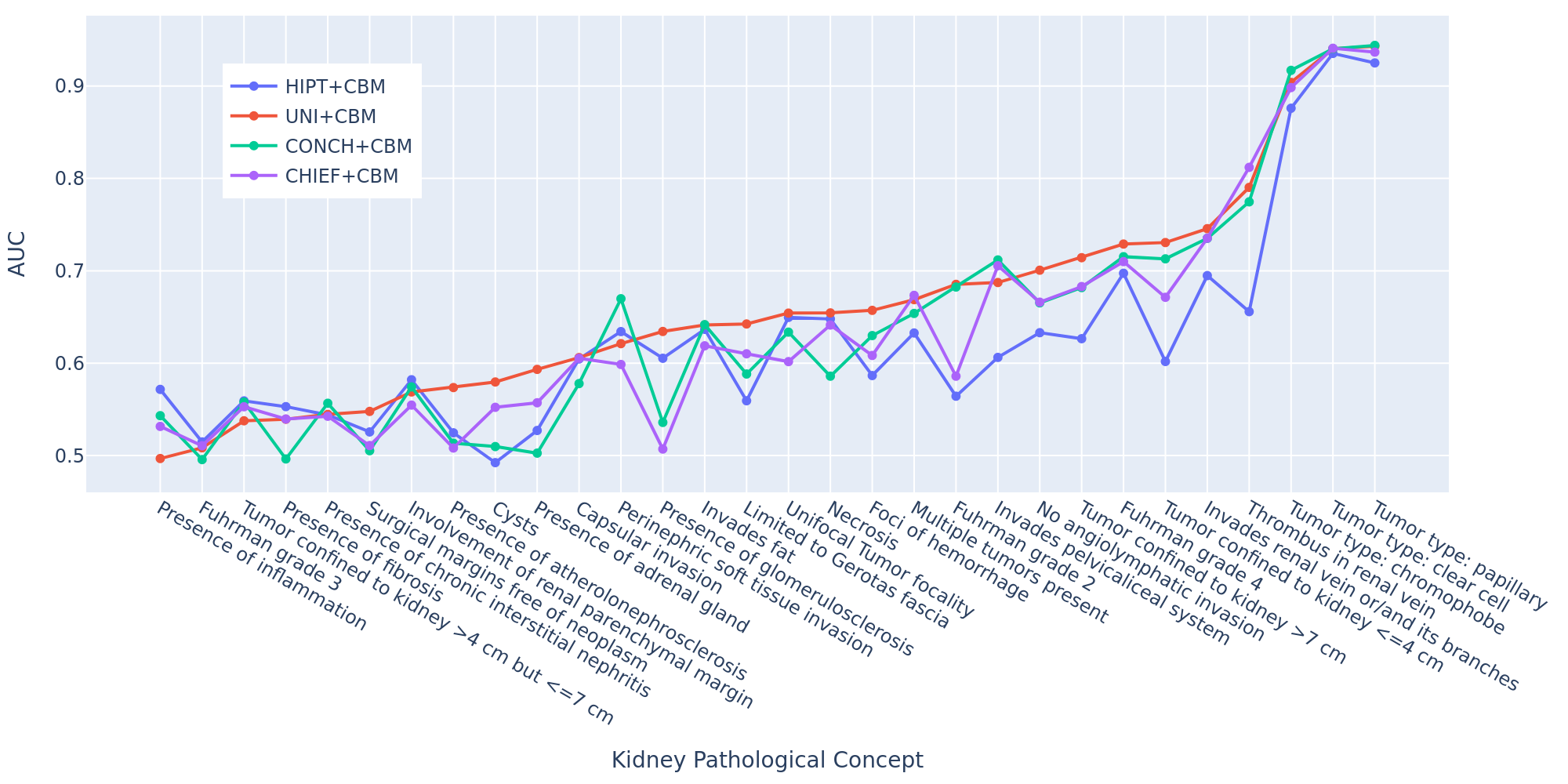}
    \caption{Benchmarking foundation models by pathological concept learning on kidney cancer.}
    \label{fig:concept learning}
\end{figure}

\textbf{Specific concept learning performance.} Figure \ref{fig:concept learning} visualizes the AUC scores across all concepts, with each line representing a specific model. Notably, the performance of cancer subtyping concepts consistently ranked among the top three in performance, which is consistent with the known differences in outcomes between kidney cancer subtypes. Additionally, tumor size-related concepts, such as "Tumor confined to kidney <=4 cm" and "Tumor confined to kidney >7 cm," demonstrated strong performance. These concepts require global information fusion rather than localized patch-level embeddings, highlighting the effectiveness of our GNN-based approach in integrating spatial information across patches to generate a comprehensive representation.
Some concepts such as "Surgical margins free of neoplasm" showed lower performance. This can be attributed to their local nature; in general, such concepts require the analysis of very specific tissue slides which may not be available in public datasets. The rarity of a concept also contributed to their lower AUC score.
There is overlap between some of the concepts, for example between those concerning grade or renal vein invasion; refining these definitions could likely improve performance.

\textbf{Cancer Subtyping.}
Table \ref{tab:tumor subtyping} presents the results for kidney cancer subtyping. Overall, all models exhibited significantly higher performance in cancer subtyping compared to other concept learning tasks. Notably, UNI + CBM and CONCH + CBM achieved comparable results, with UNI + CBM outperforming in 5 out of 12 metrics and CONCH + CBM leading in 6 metrics. Additionally, CHIEF + CBM achieved the highest performance in one metric. These findings suggest that cancer subtyping relies on a more limited subset of information within WSIs, making it less effective for assessing the feature extraction capabilities of foundation models.

\begin{table}[!tb]
    \centering
    \begin{tabular}{c|c|c|c|c|c|c|c|c|c|c|c|c}
        \hline
        \multirow{2}{*}{Model} & \multicolumn{4}{c|}{ccRCC} & \multicolumn{4}{c|}{pRCC}& \multicolumn{4}{c}{chRCC} \\
        \cline{2-13}
        &  ACC& F1& AUC& AP& ACC& F1& AUC& AP& ACC& F1& AUC& AP\\
        \hline
        HIPT+CBM & .890& .912& .935& .936& .853& .807& .925& .888& .686& .523& .876& .666\\
        UNI+CBM & \textbf{.918}& \textbf{.935}& .940& \textbf{.942}& .902& .876& .943& .930& \textbf{.791}& \textbf{.710}& .904& .764\\
        CONCH+CBM & .907& .926& .940& .941& \textbf{.907}& \textbf{.886}& \textbf{.944}& \textbf{.935}& .756& .654& \textbf{.917}& \textbf{.788}\\
        CHIEF+CBM & .906& .927& \textbf{.941}& .940& .880& .846& .937& .911& .769& .667& .898& .761\\
        \hline
    \end{tabular}
    \caption{Kidney cancer subtyping. ccRCC, pRCC, and chRCC refer to clear cell RCC, papillary RCC, and chromophobe RCC, respectively. }
    \label{tab:tumor subtyping}
\end{table}

\subsection{Survival analysis} \label{experiment:survival analysis}
In this section, we conducted kidney cancer overall survival analysis using four different settings: (1) applying CoxPH regression with binary concept labels, (2) training end-to-end models using the same loss function as CoxPH, (3) performing regression on deep features aggregated by the end-to-end models, and (4) using explainable concepts identified by CBM for regression. For CoxPH regression, we applied 
$\ell_2$ regularization to prevent overfitting and determined the optimal regularization parameter through 5-fold cross-validation on the training samples. Table \ref{tab:survival} presents the results of kidney cancer survival analysis across these four settings.

Among the end-to-end models (1–4), the CONCH-based model achieved the best C-Index and C-AUC, while the UNI-based model performed best in terms of C-IPCW. Since the end-to-end models cannot predict time-dependent survival functions, the IBS score is not applicable to them. To assess overfitting in the end-to-end models, we used aggregated features from ABMIL for CoxPH regression. Comparisons between the end-to-end models (1–4) and the two-stage models (5–8) shows that both groups delivered comparable performance, indicating no significant overfitting. Finally, when the soft logits inferred by CBM were fed into CoxPH for regression, the comparison between the ABMIL-based models (5–8) and the CBM-based models (9–12) revealed that CBM consistently improved model performance as C-Index increased and IBS decreased across all feature extractors. Moreover, while ABMIL outputs hidden features, CBM provides explainable concepts, offering valuable insights for clinicians. Notably, models 10-12, which were trained to identify 30 concepts, outperformed model 13 across all metrics. This improvement may be due to the soft logits potentially retaining image information, unlike the hard binary labels.

\begin{table}[!tb]
    \centering
    \begin{tabular}{c|c|c|c|c|c|c}
        \hline
        Model &Feature & Regression& C-Index& C-IPCW & C-AUC& IBS \\
        \hline
        1 & HIPT+ABMIL & MLP & 0.672$_{0.032}$& 0.649$_{0.050}$& 0.675$_{0.030}$& N/A  \\
        2 & UNI+ABMIL & MLP & 0.682$_{0.050}$& \textbf{0.698}$_{0.082}$& 0.694$_{0.061}$& N/A  \\
        3 & CONCH+ABMIL & MLP & 0.702$_{0.024}$& 0.671$_{0.038}$& 0.706$_{0.031}$& N/A  \\
        4 & CHIEF+ABMIL & MLP & 0.679$_{0.031}$& 0.646$_{0.025}$& 0.688$_{0.038}$& N/A  \\
        \hline
        5 & HIPT+ABMIL & CoxPH & 0.671$_{0.034}$& 0.615$_{0.019}$& 0.672$_{0.034}$& 0.192$_{0.017}$  \\
        6 & UNI+ABMIL & CoxPH & 0.682$_{0.050}$& \textit{0.693}$_{0.083}$& 0.692$_{0.059}$& 0.189$_{0.042}$  \\
        7 & CONCH+ABMIL & CoxPH & 0.694$_{0.029}$& 0.656$_{0.052}$& 0.698$_{0.034}$& 0.187$_{0.017}$  \\
        8 & CHIEF+ABMIL & CoxPH & 0.673$_{0.038}$& 0.650$_{0.040}$& 0.682$_{0.042}$& 0.212$_{0.019}$  \\
        \hline
        9 & HIPT+CBM & CoxPH & 0.678$_{0.046}$& 0.648$_{0.042}$& 0.671$_{0.052}$& 0.168$_{0.013}$  \\
        10 & UNI+CBM & CoxPH & 0.714$_{0.026}$& 0.650$_{0.039}$& 0.715$_{0.036}$& 0.162$_{0.010}$  \\
        11 & CONCH+CBM & CoxPH & \textbf{0.725}$_{0.023}$& 0.678$_{0.036}$& \textbf{0.725}$_{0.021}$& \textbf{0.159}$_{0.015}$  \\
        12 & CHIEF+CBM & CoxPH & 0.715$_{0.034}$& \textbf{0.698}$_{0.074}$& 0.711$_{0.044}$& \textbf{0.159}$_{0.010}$ \\
        \hline
        13 & 30 concepts& CoxPH& 0.707$_{0.010}$& 0.647$_{0.046}$& 0.706$_{0.010}$& 0.170$_{0.012}$  \\
        \hline
    \end{tabular}
    \caption{Kidney cancer survival analysis. Model 1-8 used deep features aggregated by ABMIL. Model 9-12 used explainable concepts aggregated by CBM. Model 13 used the binary labels of all 30 concepts.}
    \label{tab:survival}
\end{table}


\begin{figure}[!tb]
    \centering
    \includegraphics[width=1\linewidth]{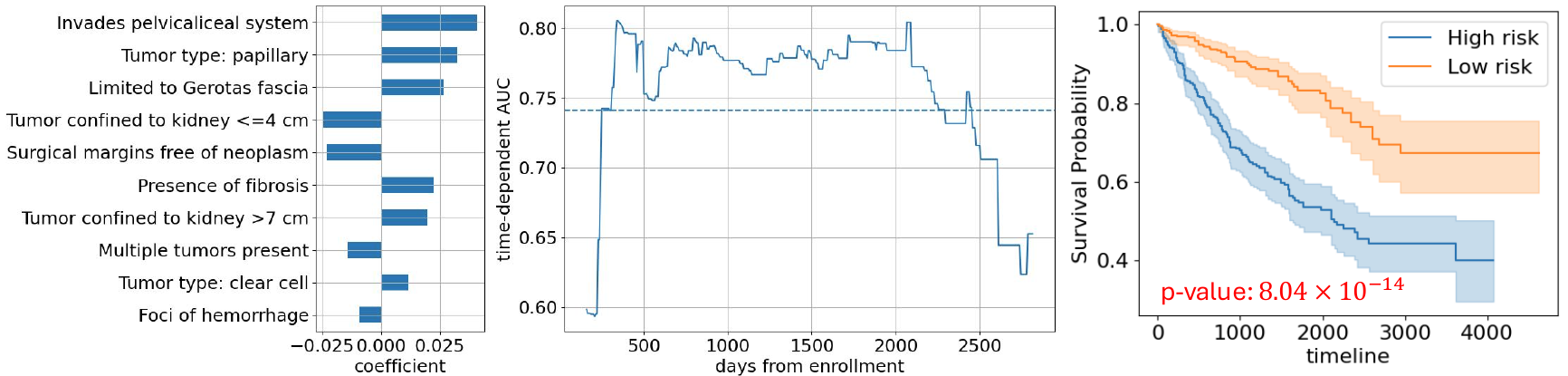}
    \caption{Kidney cancer survival analysis. The left shows the coefficients of top 10 high risk factors leading to mortality. The middle shows the AUC at different time points. The right shows the survival curves of high- and low-risk groups.}
    \label{fig:survival}
\end{figure}

\subsection{Explainability and Fairness}
In this section, we examine the explainability and fairness of our pathological concept learning for survival analysis, using model 11 from Table \ref{tab:survival}. To assess the importance of each concept in predicting survival, we analyzed the coefficients of the top 10 high-risk factors contributing to cancer mortality, as shown in the left figure of Fig. \ref{fig:survival}. Notably, "Invades pelvicaliceal system" emerged as the most significant factor associated with increased mortality, which aligns with previous studies highlighting its link to poor survival outcomes \cite{renal_reports}. Conversely, "Tumor confined to kidney <=4cm" was identified as the most significant factor decreasing mortality, as smaller tumors tend to be less aggressive. 

To evaluate the reliability of survival predictions at different time points, we analyzed the time-dependent AUC in the middle figure. The results indicate that overall survival predictions are more reliable for survival times ranging from approximately 300 to 2200 days.

To demonstrate the effectiveness of discriminating between low- and high-risk groups, we separated the samples based on their risk scores by comparing them to the mean risk score. The survival functions for these groups, fitted using the Kaplan-Meier estimator, are shown in the right figure. The log-rank test yielded a p-value below 0.005, confirming the effectiveness of pathological concept learning in distinguishing between low- and high-risk groups.

\begin{figure}[!tb]
    \centering
    \includegraphics[width=1\linewidth]{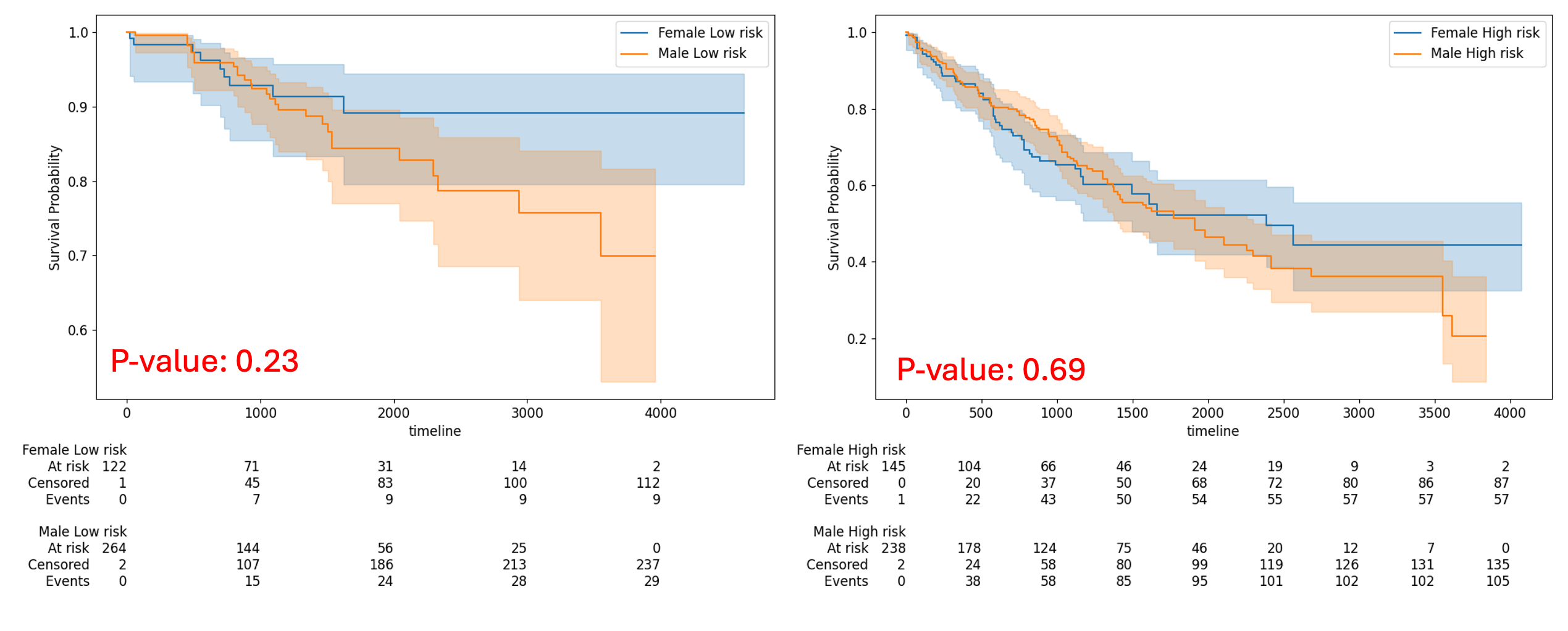}
    \caption{Evaluation of fairness in terms of gender. The left(right) shows the survival curves of female and male low-risk(high-risk) groups. }
    \label{fig:faireness2gender}
\end{figure}

\begin{figure}[!tb]
    \centering
    \includegraphics[width=1\linewidth]{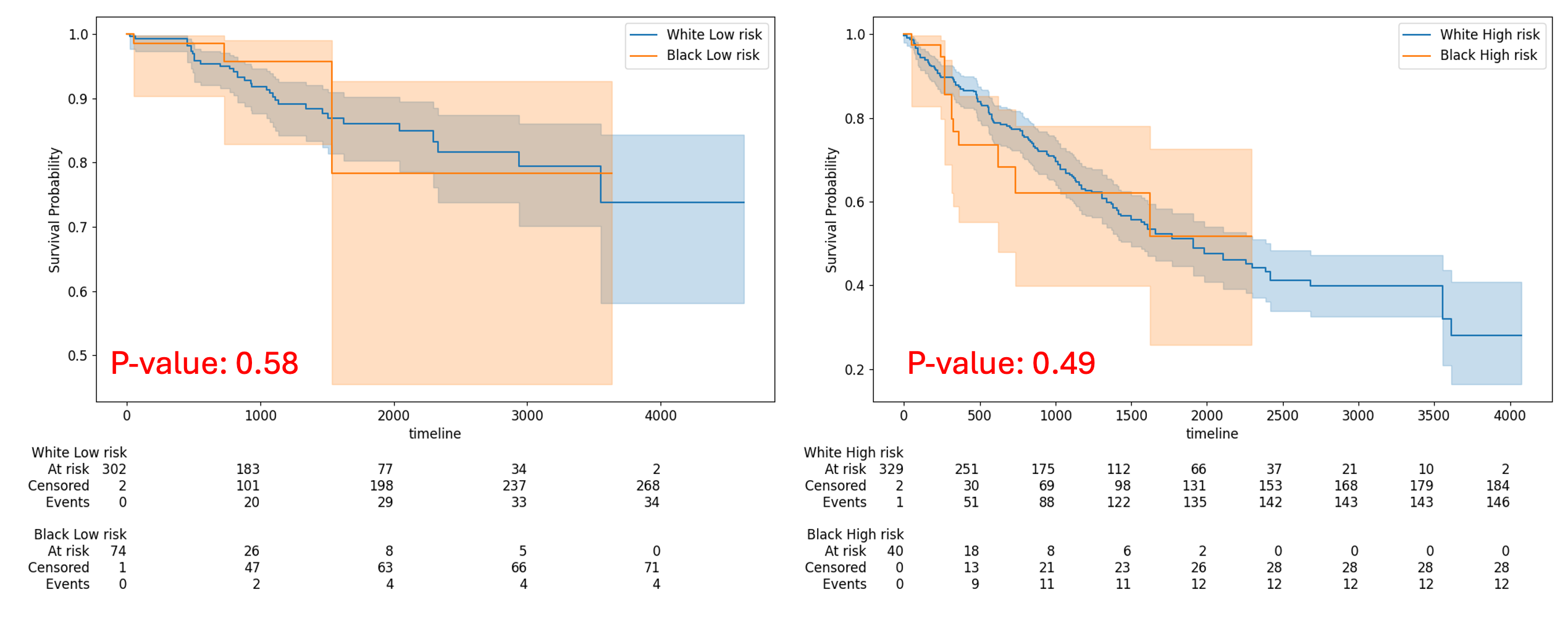}
    \caption{Evaluation of fairness in terms of race. The left(right) shows the survival curves of white and black low-risk(high-risk) groups.}
    \label{fig:faireness2race}
\end{figure}

To evaluate the fairness in terms of gender, we split the low- or high-risk patients into female and male groups, and conducted the log-rank tests, as shown in Fig. \ref{fig:faireness2gender}. The left figure shows that the low-risk group consists of 123 female patients and 266 male patients. The p-value of log-rank test is 0.23. The right fiture shows that the high-risk group consists of 146 female patients and 240 male patients. The p-value of log-rank test is 0.69. Both p-values are larger than 0.005, demonstrating the CBM-based risk identification method has no bias to gender. 

To evaluate the fairness in terms of race, we included white and black patients while excluded asian patients and others, since the number of asian patients (only 9) and others (only 15) is insufficient for analysis. Then, we split the low- or high-risk patients into white and black groups, and conducted the log-rank tests, as shown in Fig. \ref{fig:faireness2race}. The left figure shows that the low-risk group consists of 304 while patients and 75 black patients. The p-value of log-rank test is 0.58. The right figure shows that the high-risk group consists of 332 white patients and 40 black patients. The p-value of log-rank test is 0.49. These results demonstrate the CBM-based risk identification method has no bias to race.

\section{Conclusion}
In this work, we developed a pathological concept learning framework for kidney cancer, and demonstrated its explainability and fairness for survival analysis by identifying interpretable high-risk factors related to survival and validating its effectiveness in discriminating low- and high-risk groups.

\noindent\textbf{Prospect of application:}  These strengths highlight its potential in subtype-specific analyses and prediction of kidney cancer relapses to provide additional clinical insights in patient management and surveillance planning.

    

\begin{credits}
\subsubsection{\ackname} This work was supported by the Cancer Research UK Cambridge Centre [CTRQQR-2021\textbackslash 100012; and C9685/A25117], The Mark Foundation for Cancer Research [RG95043],  NIHR Cambridge Biomedical Research Centre (NIHR203312), and the EPSRC Tier-2 capital grant [EP/P020259/1]. M.C.O. was supported by the Joseph Mitchell Cancer Research Fund, the Academy of Medical Sciences [G117526] and NIHR [NIHR206092]. The results shown here are in whole or part based upon data generated by the TCGA Research Network: http://cancergenome.nih.gov/.

\subsubsection{\discintname} The authors have no competing interests to declare that are
relevant to the content of this article.
\end{credits}

%
%
%
%

\end{document}